\def\year{2017}\relax
\documentclass[letterpaper]{article}
\usepackage{aaai17}
\usepackage{times}
\usepackage{helvet}
\usepackage{courier}
\usepackage{dirtytalk}
\usepackage{times}
\usepackage{url}
\usepackage{latexsym}
\usepackage{graphicx}
\usepackage{xcolor}
\usepackage[utf8]{inputenc}
\usepackage{enumitem}
\usepackage{caption}
\usepackage{subcaption}
\usepackage{epstopdf}
\usepackage{multirow}
\usepackage[title,titletoc,toc]{appendix}
\usepackage{amsmath}
\usepackage{amsfonts}

\newcommand{\quotes}[1]{``#1''}
\newcommand{\YA}{Yahoo!\ Answers }

\newcommand{\X}{\mathbf{x}}

\pdfoutput=1
\begin{document}

\title{Community Question Answering Platforms vs. Twitter for Predicting Characteristics of Urban Neighbourhoods}
 \author{Marzieh Saeidi \\ University College London \\m.saeidi@cs.ucl.ac.uk\And Licia Capra \\ University College London \\l.capra@ucl.ac.uk \AND Alessandro Venerandi \\University College London \\alessandro.venerandi.12@ucl.ac.uk \And Sebastian Riedel \\ University College London \\s.riedel@cs.ucl.ac.uk }
 
\maketitle
\begin{abstract}
In this paper, we investigate whether text from a Community Question Answering (QA) platform can be used to predict and describe real-world attributes. We experiment with predicting a wide range of $62$ demographic attributes for neighbourhoods of London. We use the text from QA platform of \YA and compare our results to the ones obtained from Twitter microblogs. Outcomes show that the correlation between the predicted demographic attributes using text from \YA discussions and the observed demographic attributes can reach an average Pearson correlation coefficient of $\rho=0.54$, slightly higher than the predictions obtained using Twitter data. Our qualitative analysis indicates that there is semantic relatedness between the highest correlated terms extracted from both datasets and their relative demographic attributes. Furthermore, the correlations highlight the different natures of the information contained in \YA and Twitter. While the former seems to offer a more encyclopedic content, the latter provides information related to current sociocultural aspects. 

\end{abstract}

\section{Introduction}
Recent years have seen a huge boom in the number of different social media platforms available to users. People are increasingly using these platforms to voice their opinions or let others know about their whereabouts and activities. Each of these platforms has its own characteristics and is used for different purposes. The availability of a huge amount of data from many social media platforms has inspired researchers to study the relation between the data generated through the use of these platforms and real-world attributes. 
 
Many recent studies in this field are particularly inspired by the availability of text-based social media platforms such as blogs and Twitter. Text from Twitter microblogs, in particular, has been widely used as data source to make predictions in many domains. For example, box-office revenues are predicted using text from Twitter~\cite{asur2010predicting}. Twitter data has also been used to find correlations between the mood stated in tweets and the value of Dow Jones Industrial Average (DJIA)~\cite{bollen2011twitter}. 

Predicting demographics of individual users using their language on social media platforms, especially Twitter, has been the focus of many research works: text from blogs and on-line forum posts are utilised to predict user's age through the analysis of linguistic features. Results show that the age of users can be predicted where the predicted and observed values reach a Pearson correlation coefficient of almost $0.7$. Sociolinguistic associations using geo-tagged Twitter data have been discovered~\cite{eisenstein2011discovering} and the results indicate that the demographic information of users such as first language, race, and ethnicity can be predicted by using text from Twitter with a correlation up to $0.3$. Other research shows that users' income can also be predicted using tweets with a good prediction accuracy~\cite{preoctiuc2015analysis}. Text from Twitter microblogs has also been used to discover the relation between the language of users and the deprivation index of neighbourhoods. The collective sentiment extracted from the tweets of users has been shown~\cite{quercia2012tracking} to have significant correlation ($0.35$) with the deprivation index of the communities the users belong to. 

Data generated on QA platforms have not been used in the past for predicting real-world attributes. Most research work that utilise QA data aim to increase the performance of such platforms in analysing question quality~\cite{li2012analyzing}, predicting the best answers~\cite{liu2010predicting,tian2013predicting} or the best responder~\cite{zhao2012predicting}.

In this paper, we use the text from the discussions on the QA platform of \YA about neighbourhoods of London to show that the QA text can be used to predict the demographic attributes of the population of those neighbourhoods. We compare the performance of \YA data to the performance of data from Twitter, a platform that has been widely used for predicting many real-world attributes. Unlike many current works that focus on predicting one or few selected attributes (e.g. deprivation, race or income) using social media data, we study a wide range of $62$ demographic attributes. Furthermore, we test whether terms extracted from both \YA and Twitter are semantically related to these attributes and provide examples of sociocultural profiles of neighbourhoods through the interpretation of the coefficients of the predictive models. 

\begin{table*}[ht]
\small
\caption{Examples of \YA discussions and Twitter microblogs about neighbourhoods that contain the term \say{Jewish}.}
\centering
\begin{tabular}{|p{16cm}|}
\hline
\textbf{\YA}\\
\hline
\textbf{Q:} Where can i find a jewish shop in london?\\
\textbf{A:} The main \textbf{Jewish} Communities in London are \underline{Stamford Hill} and \underline{Golders Green}, plus \underline{Hendon} and \underline{Edgeware}. All have many Kosher and Judaica stores on their high streets.\\ 
\\
\textbf{Q:} Jewish neighborhoods in London?\\
\textbf{A:} The largest is in \underline{Gants Hill}. They are predominantly Reformist Jews. Then you have the largest Hasidic \textbf{Jewish} Community in Europe in \underline{Stamford Hill}. Then there is a large Orthodox \textbf{Jewish} Community in \underline{Hendon}, and around 14\% of \underline{Swiss Cottage} is \textbf{Jewish}. \\
\hline
\textbf{Twitter}\\
\hline
- Meanwhile in Camden. @ \textbf{Jewish} Museum London  [tweeted from \underline{Camden}]\\
- Challah makes me happy. Braided, proofed and egg washed \#Shabbatshalom \#dough \#sesame \#\textbf{Jewish}food… [tweeted from \underline{East Finchley}]\\
- \textbf{Jewish} crouton crack. For when you just need that boost \#osem \#mondaymorning \#whoneedschickensoup [tweeted from \underline{Golders Green}]\\
\hline
\end{tabular}
\label{table_examples_ya_twitter_descriptive}
\end{table*}

The contributions of this paper can be summarised as follows:
\begin{itemize}
	\item We show that text from QA discussions can be used to predict real-world attributes such as demographic attributes of the population of neighbourhoods with a performance comparable to Twitter data.
	\item Our analysis highlights the differences between data from a QA platform and Twitter: while QA data offers a more encyclopedic content, the latter provides information related to current sociocultural aspects.
\end{itemize}

 
\section{Datasets}
\paragraph{\YA.} 
Community QA platforms help users to obtain information from a community -- a user can post questions which may then be answered by other users. Discussions can be in depth and in length but the interactions are not spontaneous. The time frame in which users take part in a discussion thread can vary from one day to several days or even to months. Moreover, QA platforms, unlike Twitter and some of the other social media platforms, are not location-based. \YA is one of the few QA platforms that have emerged in the past decade. Discussions in \YA are not domain specific and can cover a broad range of topics. 

\paragraph{Twitter.} 
Twitter is an on-line microblogging social network where users can post and read short 140-character messages. Twitter is used mostly to share views, opinions and news in real-time. Unlike QA platforms, Twitter is ubiquitous. While some users use Twitter to get updates on the news and their social circle, many others use it as part of their daily routine to talk about their thoughts, whereabouts, activities or sometimes just to share what is going on in their lives. This can be because of the strong tie that there exists between Twitter and smartphones which are nowadays at the centre of many people's lives. 
For all these reasons, huge amount of data is constantly being created on this platform. Additionally, Twitter is a location-based platform where people can tag their locations while blogging their tweets. 

\paragraph{\YA vs. Twitter.}
When it comes to neighbourhoods, \YA have been used by many users to ask or to answer questions about different characteristics of many neighbourhoods. While people may not use Twitter in the same way, they may log their tweets while being in different neighbourhoods. In this paper, we investigate the extent to which the discussion on \YA platforms about neighbourhoods and the microblogs that are logged from different neighbourhoods can reflect characteristics of those neighbourhoods. Table~\ref{table_examples_ya_twitter_descriptive} shows examples of \YA discussions that contain the names of London neighbourhoods and the term \quotes{Jewish}. The table also shows examples of tweets that have been blogged from London neighbourhoods (neighbourhood's name in brackets) and contain the term \quotes{Jewish}. These examples show some of the differences between the discussions that can be found on \YA and on Twitter. As we can see, the QA discussions are focused on a topic, i.e. \textit{Jewish} neighbourhoods in London. The answers provide explicit information on neighbourhoods that have a high population of Jewish. On the other hand, microblogs of Twitter do not focus on providing explicit information about the neighbourhoods. But they contain information on user's activities or observations (e.g. Jewish Museum, Jewish food) while in different locations. These activities can implicitly indicate that Jewish communities inhibit the neighbourhoods that the user is blogging from.  

\paragraph{Population Demographic Data.} 
Population demographic data is taken from the UK census provided by the Office for National Statistics.~\footnote{\url{http://www.ons.gov.uk/}} Census surveys in the UK are repeated every $10$ years and were last conducted in 2011. Census data is provided for specific geographical units that are created solely for the purpose of census data collection. These are called Lower Layer Super Output areas (LSOAs) and are identified through an alphanumeric ID. Greater London is divided into $4,835$ LSOAs. These are not necessarily equal in size as they have been designed to have a population of around $1,500$. Census data provides statistical information on a wide range of categories such as the average house price in a LSOA, population count (or percent) of a religion or an ethnic background, income level, etc. Each category can be subdivided into further attributes. For instance, the category religion contains the attributes Muslim, Christian, Jewish, Hindu, etc.

\section{Method}
\subsection{Spatial Unit of Analysis}
The spatial unit of analysis chosen for this work is the neighbourhood. This is identified with a unique name (e.g., Camden) and people normally use this name in QA discussions to refer to specific neighbourhoods.
A list of neighbourhoods for London is extracted from the GeoNames gazetteer\footnote{\url{http://www.geonames.org/}}, a dataset containing names of geographic places including place names. For each neighbourhood, GeoNames provides its name and a set of geographic coordinates (i.e., latitude and longitude) which roughly represents its centre. Note that geographical boundaries are not provided. GeoNames contains $589$ neighbourhoods that fall within the boundaries of the Greater London metropolitan area. In the remainder of the paper, we use the terms \quotes{neighbourhood} or \quotes{area} to refer to our spatial unit of analysis.

\subsection{Pre-processing, Filtering, and Spatial Aggregation}

\subsubsection{\YA Data.} 
\begin{figure}[ht]
    \centering
    \includegraphics[width=0.45\textwidth]{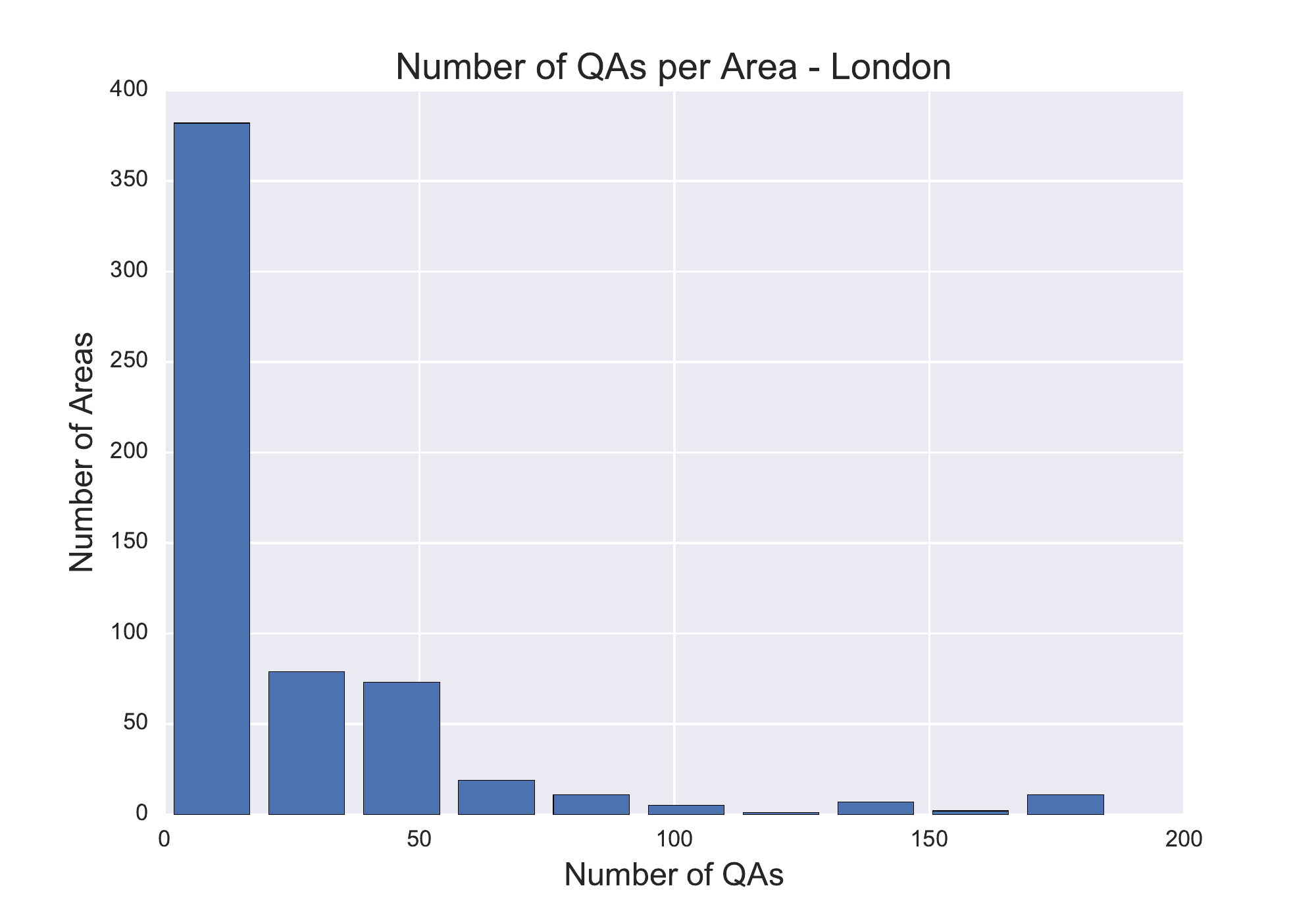}
    \caption{Histogram of the number of QAs per each London neighbourhood.}
    \label{fig_hist_area_ya_count}
\end{figure}

We collect questions and answers (QAs) from \YA using its public API.\footnote{\url{https://developer.yahoo.com/answers/}} For each neighbourhood, the query consists of the name of the neighbourhood together with the keywords \say{London} and \say{area}. This is to prevent obtaining irrelevant QAs for ambiguous entity names such as Victoria. For each neighbourhood, we then take all the QAs that are returned by the API. Each QA consists of a title and a content which is an elaboration on the title. This is followed by a number of answers. In total, we collect $12,947$ QAs across all London neighbourhoods. These QAs span over the last 5 years. It is common for users to discuss characteristics of several neighbourhoods in the same QA thread. This means that the same QA can be assigned to more than one neighbourhood.
Figure~\ref{fig_hist_area_ya_count} shows the histogram of the number of QAs for each neighbourhood. As the figure shows, the majority of areas have less than $100$ QAs with some areas having less than $10$. Only few areas have over $100$ QAs. 

For each neighbourhood, we create one single document by combining all the QA discussions that have been retrieved using the name of such neighbourhood. This document may or may not contain names of other neighbourhoods. We split each document into sentences and remove those neighbourhoods containing less than $40$ sentences. 

We then remove URLs from each document. The document is then converted to tokens and stop words are removed. All the tokens in all the documents are then stemmed. The goal of stemming is to reduce the different grammatical forms of a word to a common base form. Stemming is a special case of text normalisation. For example, a stemmer will transform the word \quotes{presumably} to \quotes{presum} and \quotes{provision} to \quotes{provis}. To keep the most frequent words, we remove any token that has appeared less than $5$ times in less than $5$ unique QAs. This leaves us with $8$k distinct tokens.

\subsubsection{Twitter Data.}
To collect data from Twitter, we use the geographical bounding box of London, defined by the northwest and southeast points of the Greater London region. We then use this bounding box to obtain the tweets that are geotagged and are created within this box through the official Twitter API.\footnote{\url{https://dev.twitter.com/streaming/}} We stream Twitter data for $6$ months between December 2015 and July 2016. At the end, we have around $2,000,000$ tweets in our dataset. 

\begin{figure}[ht]
    \centering
    \includegraphics[width=0.45\textwidth]{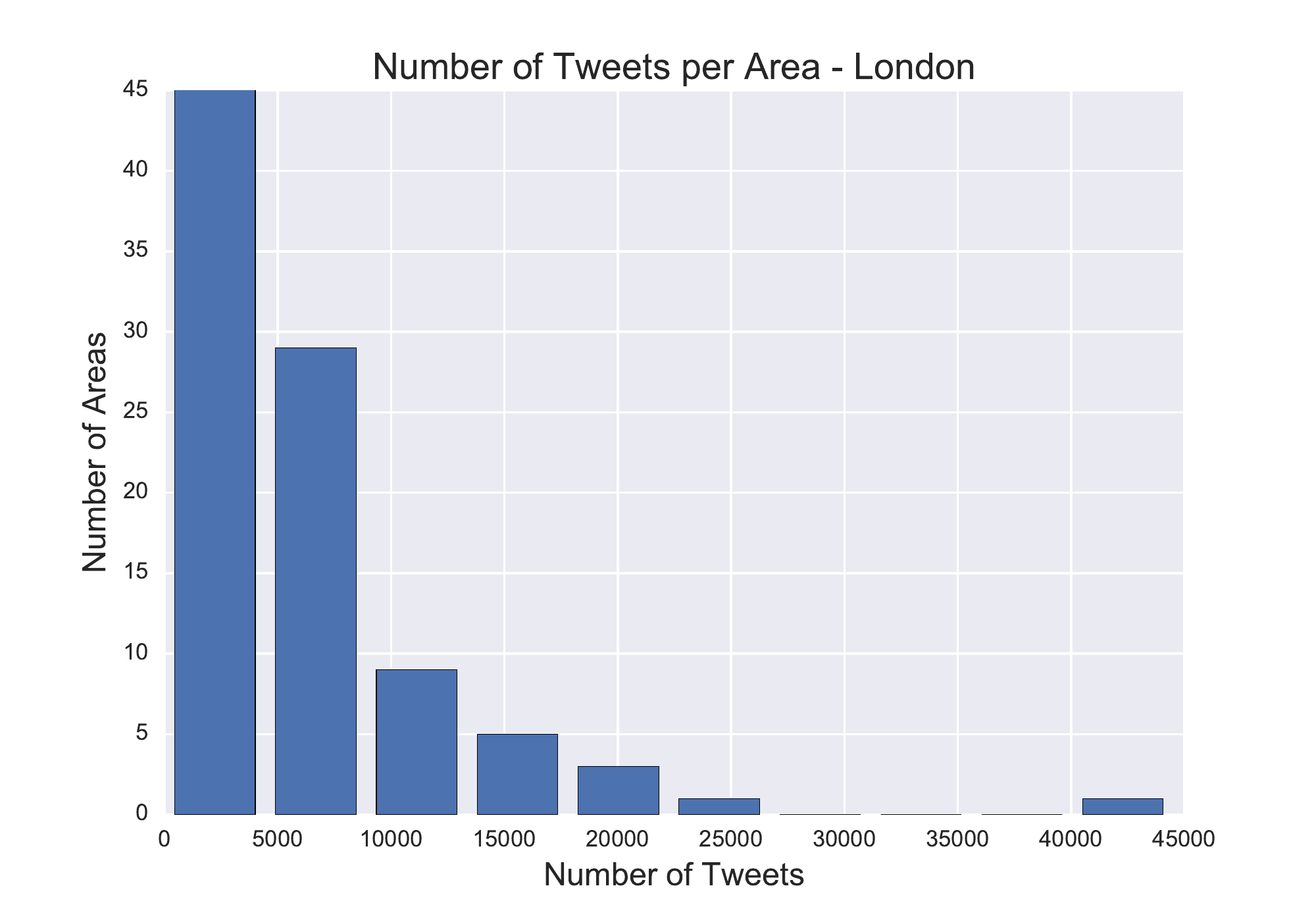}
    \caption{Histogram of the number of tweets per each London neighbourhood.}
    \label{figure_tweet_counts_hist}
\end{figure}

To assign tweets to different neighbourhoods, for each tweet, we calculate the distance between the location that it was blogged from and the centre points of all the neighbourhoods in our dataset. Note that the centre point for each neighbourhood is provided in the gazetteer. We then assign the tweet to the closest neighbourhood that is not further than $1$ km from the tweet's geolocation. At the end of this process, we have a collection of tweets per each neighbourhood and we combine them to create a single document. Figure~\ref{figure_tweet_counts_hist} shows the number of tweets per each neighbourhood. As we can see, the majority of neighbourhoods have less than $1000$ tweets. 

We remove all the target words (words starting with @) from the documents. The pre-processing is then similar to the QA documents. At the end of this process, we obtain $17$k distinct frequent tokens for the Twitter corpus.

\subsubsection{Population Demographic Data.} 
\begin{figure}[ht]
    \centering
        \centering
        \includegraphics[width=0.49\textwidth]{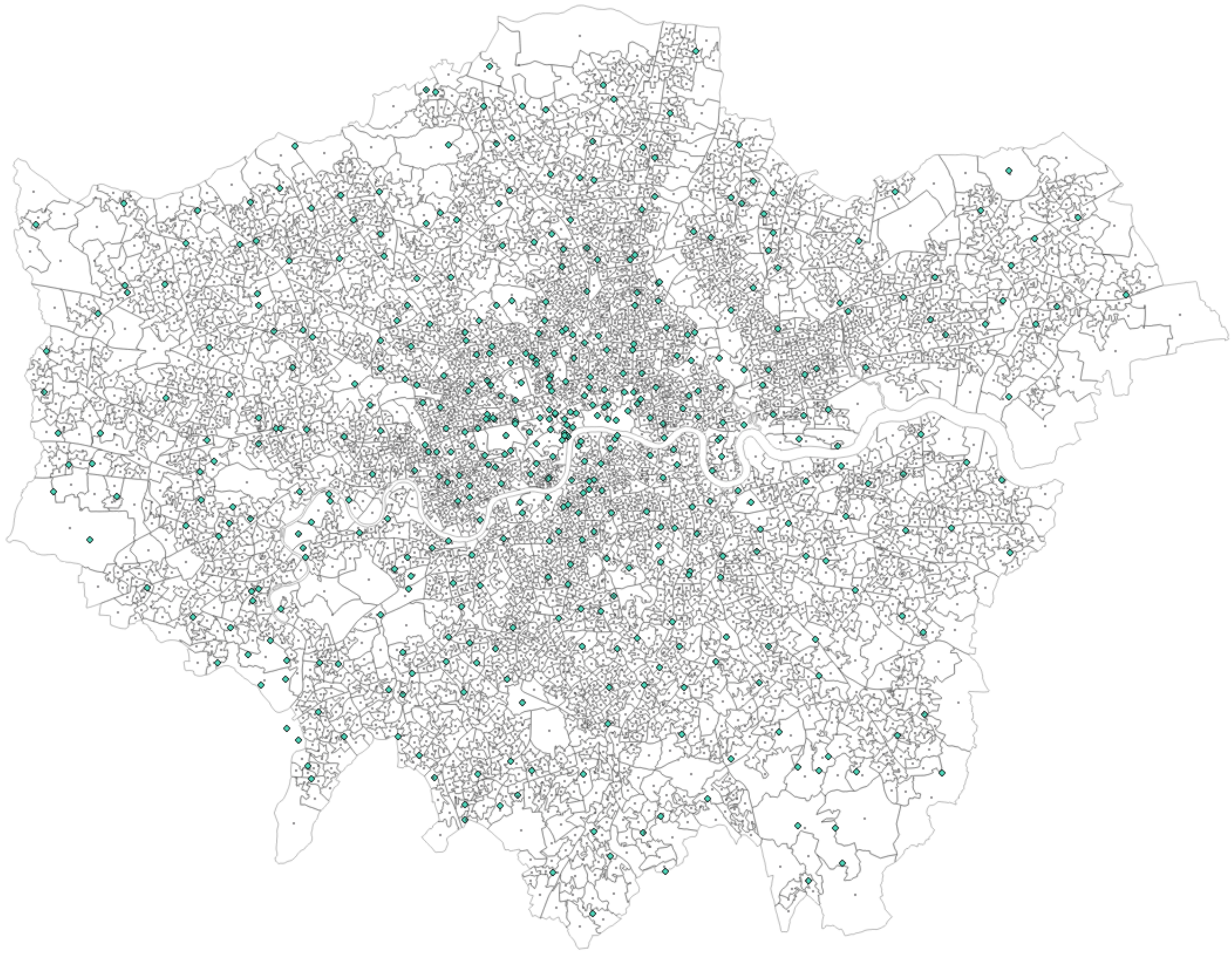}
    \caption{The geographic relation between LSOAs and neighbourhoods identified through the gazetteer. LSOAs are geographical shapes and their centroids are marked with dark dots. Neighbourhoods are marked with green circles.}
    \label{fig_lsoa_area_mapping}
\end{figure}

As we previously explained, each attribute in census data is assigned to spatial units called LSOAs. However, these units do not geographically match our units of analysis which are the neighbourhoods defined trough the gazetteer. A map showing the spatial mismatch is presented in Figure~\ref{fig_lsoa_area_mapping}. To aggregate the data contained in the LSOAs at the neighbourhood level, we use the following approach.

Often, when people talk about a neighbourhood, they refer to the area around its centre point. Therefore, the information provided for neighbourhoods in QA discussions should be very related to this geographic point. To keep this level of local information, for each demographic attribute, we assign only the values of the nearby LSOAs to the respective neighbourhood. To do this, we calculate the distance between each neighbourhood and all the LSOAs in London. The distance is calculated between the coordinates of a neighbourhood and the coordinates of each LSOA's centroid. For each neighbourhood, we then select the $10$ closest LSOAs that are not further than one kilometre away. The value of each demographic attribute for each neighbourhood is then computed by averaging the values associated with the LSOAs assigned to it. We apply this procedure to all the demographic attributes. 

\subsection{Document Representation}
A very popular method for representing a document using its words is the tf-idf approach~\cite{salton1983extended}. Tf-idf is short for term frequency-inverse document frequency where tf indicates the frequency of a term in the document and idf is a function of the number of documents that a terms has appeared in. In a tf-idf representation, the order of the words in the document is not preserved. For each term in a document, the tf-idf value is calculated as below:

\begin{equation}
\small
\text{tf-idf} (d,t) = \frac{\text{tf}\ (d,t)}{\log(\frac{\text{Total number of documents}}{\text{Number of documents containing the  term}\ t })}	
\end{equation}
To discount the bias for areas that have a high number of QAs or tweets, we normalise tf values by the length of each document as below. The length of a document is defined by the number of its tokens (non-distinctive words). 

\begin{equation}
\small
\text{Normalised\ tf} (d,t) = \frac{\text{Frequency of Term t in Document d}}{\text{Number of Tokens in Document d}}
\end{equation}

\subsection{Correlation}
To investigate the extent to which the text obtained from the two platforms of \YA and Twitter reflect the true attributes of neighbourhoods, we first study whether there are significant, strong and meaningful correlations between the terms present in each corpus and the many neighbourhood attributes through the Pearson correlation coefficient $\rho$. For each term in each corpus, we calculate the correlation between the term and all the selected demographic attributes. To do so, for each term, we define a vector with the dimension of the number of neighbourhoods. The value of each cell in this vector represents the normalised tf-idf value of the term for the corresponding neighbourhood. For each demographic attribute, we also define a vector with the dimension of the number of neighbourhoods. Each cell represents the value for the demographic attribute of the corresponding neighbourhood. We then calculate the Pearson correlation coefficient ($\rho$) between these two vectors to measure the strength of the association between each term and each attribute. 

Since we perform many correlation tests simultaneously, we need to correct the significance values (p-values) for multiple testing. We do so by implementing the Bonferroni correction, a multiple-comparison p-value correction, which is used when several dependent or independent statistical tests are being performed simultaneously. The Bonferroni adjustment ensures an upper bound for the probability of having an erroneous significant result among all the tests. All the p-values showed in this paper are adjusted through the use of the Bonferroni correction. 

\subsection{Prediction}
We investigate how well the demographic attributes can be predicted by using using Yahoo! Ansewrs and Twitter data. We define the task of predicting a continuous-valued demographic attribute for unseen neighbourhoods as a regression task given their normalised tf-idf document representation. A separate regression task is defined for each demographic attribute. We choose linear regression for the prediction tasks as it has been widely used for predictions from text in the literature~\cite{foster2013featurizing,joshi2010movie}. 

Due to the high number of features (size of vocabulary) and a small number of training points, over-fitting can occur. To avoid this issue, we use elastic net regularisation, a technique that combines the regularisation of the ridge and lasso regressions. The parameters $\theta$ are estimated by minimising the following loss function. Here, $y_i$ is the value of an attribute for the $i$-th neighbourhood, vector $\X_i$ is its document representation and $N$ is the number of neighbourhoods in the training set. 
\begin{equation}\label{eq_regression_linear_reg}
\mathfrak{L} = \frac{1}{N} \sum_{i=1}^N (y_i- \X_{i}^T \theta)^2 + \lambda_1 ||\mathbf{\theta}|| + \lambda_2 ||\mathbf{\theta}||^2
\end{equation}

\subsubsection{Evaluation.}
To measure the performance of a regression model, residual-based methods such as mean squared error are commonly used. Ranking metrics such as Pearson correlation coefficient have also been used in the literature~\cite{preoctiuc2015studying,eisenstein2011discovering}. Using a ranking measure has some advantages compared to a residual-based measure. First, ranking evaluation is more robust against extreme outliers compared to an additive residual-based evaluation measure. Second, ranking metrics are more interpretable than measures such as mean squared error~\cite{rosset2005ranking}. We thus use this method for evaluating the performance of the regression models in this work.

As further performance check, we apply a $10$ folds cross-validation to each regression task. In each fold, we use $75\%$ of the data for training and the remaining $25\%$ for validation. At the end, we report the average performance over all folds together with the standard deviation. For each demographic attribute, i.e. target value, training and validation sets are sampled using Stratified Sampling. This is a sampling method from a population, when sub-populations within this population vary. For instance, in London, there are areas with very high or very low deprivation. In these cases, it is advantageous to sample each sub-population independently and proportionally to its size. 

\section{Results}

\paragraph{Note.}
There are many attributes across several categories in the census data. Because of space limitation, we conduct most of our experiments on a selected set of attributes. These attributes are taken from \textit{religion} (population of \textit{Jewish}\%, population of \textit{Muslim}\%, population of \textit{Hindu}\%, population of \textit{Buddhist}\%), \textit{ethnicity} (population of \textit{Black}\%, population of \textit{White}\%, and population of \textit{Asian}\% ethnicity), \textit{Price} (average house prices) and \textit{deprivation} (Index of Multiple Deprivation, IMD\footnote{\url{https://en.wikipedia.org/wiki/Multiple_deprivation_index}}). For a full breakdown of results please refer to the Appendix.

\subsection{Correlation}

\subsubsection{Number of Correlated Terms.}
The number of significantly correlated terms from both \YA and the Twitter with the selected demographic attributes are shown in Table~\ref{table_ya_twitter_tfidf_correlation_counts}. Note that the number of unique (frequent) words in Twitter ($17$k) is almost twice as in \YA ($8$k). The first column shows a demographic attribute and the second column indicates the source, i.e. \YA (Y!A for short) or Twitter. The third column (\quotes{All}) shows the total number of terms that have a significant correlation with each attribute (p-value $<0.01$). The following columns show the number of terms that have a significant correlation with the attribute with a $\rho$ in the given ranges. The last column shows the number of terms that are significantly correlated with the attribute with a negative $\rho$. The data source that has the highest number of correlated terms with each attribute is highlighted in bold. 

As the table shows, terms extracted from \YA tend to be more related, in terms of the number of correlated terms, to attributes related to religion or ethnicity compared to terms from Twitter. However, for two particular attributes (i.e., \textit{Price} and \textit{Buddhist}), the number of correlated terms from Twitter is higher than the ones from \YA. These results collectively suggest that there is a wealth of terms, both in \YA and Twitter, which can be used to predict the population demographics. 

\begin{table}[ht]
\small
\centering
\caption{Number of significantly correlated terms (p-value $<0.01$) from both \YA (\say{Y!\ A}) and Twitter.}
\label{table_ya_twitter_tfidf_correlation_counts}
\begin{tabular}{p{1.2cm}|p{0.85cm}|p{0.68cm}|p{0.6cm}|p{0.9cm} |p{0.9cm} |l}
Attribute & Source & All & $>$0.4 & [0.3,0.4] & [0.2,0.3] & $<0$\\
\hline
\multirow{2}{*}{IMD} & \textbf{Y!\ A} & \textbf{115} & \textbf{1} & \textbf{48} & \textbf{66} & 0\\
 & Twitter & 17 & 0 & 10 & 7 & 0\\ 
\hline
\multirow{2}{*}{Price} & Y! A & 50 & 2 & 36 & 12 & 0\\ 
 & \textbf{Twitter} & \textbf{1120} & \textbf{312} & \textbf{533} & \textbf{275} & 0\\ 
\hline
\multirow{2}{*}{Jewish\%} & \textbf{Y!\ A} & \textbf{48} & \textbf{7} & \textbf{31} & \textbf{10} & 0\\ 
 & Twitter &  6 & 0 & 5 & 1 & 0\\ 
\hline
\multirow{2}{*}{Muslim\%} & \textbf{Y! A} & \textbf{87} & 0 & \textbf{59} & \textbf{28} & 0\\ 
 & Twitter  & 13 & \textbf{1} & 8 & 4 & 0\\ 
\hline
\multirow{2}{*}{Hindu\%} & \textbf{Y!\ A} & \textbf{8} & \textbf{2} & 3 & \textbf{3} & 0\\ 
 & Twitter &  5 & 0 & 3 & 2 & 0\\ 
\hline
\multirow{2}{*}{Buddhist\%} & Y! A & 1 & 0 & 1 & 0 & 0 \\ 
 & \textbf{Twitter} & \textbf{934} & \textbf{18} & \textbf{728} & \textbf{188} & 0\\ 
\hline
\multirow{2}{*}{Black\%} & \textbf{Y!\ A} & \textbf{114} & \textbf{4} & \textbf{59} & \textbf{51} & 0\\ 
 & Twitter & 2 & 0 & 2 & 0 & 0\\ 
\hline
\multirow{2}{*}{White\%}&\textbf{Y!\ A} & \textbf{8} & 0 & 0 & 0 & \textbf{8}\\ 
 & Twitter & 0 & 0 & 0 & 0 & 0\\ 
\hline
\multirow{2}{*}{Asian\%} & \textbf{Y!\ A} & \textbf{6} & 0 & \textbf{3} & \textbf{3} & 0\\ 
 & Twitter & 1 & 0 & 1 & 0 & 0\\ 
\end{tabular}
\end{table}

\subsubsection{Semantic Relatedness.}
In this section, we observe whether the correlations between terms and attributes are semantically meaningful. Due to the limited space, we select three attributes and their relative top correlated terms extracted from \YA (Table~\ref{table_ya_tfidf_correlation}) and Twitter (Table~\ref{table_twitter_tfidf_correlation}). We choose the attributes \textit{Price} and \textit{IMD} as they show the highest number of correlated terms for both sources. For each source, we then choose one more attribute that has the highest number of strongly correlated terms ($\rho>0.4$), i.e. Jewish\% for Yahoo! Anwsers and Buddhist\% for Twitter.
\begin{table}[ht]
\small
\centering
\caption{Terms from \YA with the highest Pearson correlation coefficients for the selected demographic attributes. Correlations are statistically significant (p-value $< 0.001$).}
\label{table_ya_tfidf_correlation}
\begin{tabular}{|l|r|l|r|l|r|}
\hline
 \multicolumn{2}{ c |}{\textbf{Jewish\%}} & \multicolumn{2}{ c |}{\textbf{(High) Price}} & \multicolumn{2}{ c |}{\textbf{Deprivation}} \\
\hline
Term & $\mathbf{\rho}$  & Term & $\mathbf{\rho}$ & Term & $\mathbf{\rho}$ \\
\hline
 \textbf{matzo} 	&	0.45	&	\textbf{townhouse} 	&	0.4	&	 \textbf{hurt} 		&	 0.4  \\
 \textbf{harmony} 	&	0.45	&	 		\textbf{fortune} 	&	0.39	&	 \textbf{poverty} 	&	 0.36 \\
 \textbf{jewish} 	&	0.41	&	\textbf{qatar} 		&	0.39	&	 \textbf{drug} 		&	 0.36 \\
  \textbf{jew} 	&	0.41	&	 	\textbf{diplomat} 	&	0.39	&	 \textbf{cockney} 	&	 0.35 \\
 unfairly 	&	0.42	&	 		\textbf{exclusive} 	&	0.37	&	 \textbf{victim} 	&	 0.35 \\
 flyover 	&	0.41	&	 		hectic 				&	0.36	& 	 \textbf{mug} 		&	 0.34 \\
 \textbf{ark} 	&	0.38	&	 		\textbf{desirable} 	&	0.35	&	 \textbf{trouble} 	&	 0.34 \\
 straw 	&	0.38	&	 		\textbf{celeb} 		&	0.34&	 \textbf{notorious} &	 0.34 \\
 arab 	&	0.39	&			\textbf{aristocratic} &	0.33&	 \textbf{rundown} 	&	 0.33 \\
 \textbf{kosher} 	&	0.32	&	\textbf{fashionable} &	0.32	&	 \textbf{slum} 		&	 0.32 \\
 \hline
\end{tabular}
\end{table}

\begin{table}
\small
\centering
\caption{Terms from Twitter with the highest Pearson correlation coefficients for the selected demographic attributes. Correlations are statistically significant (p-value $< 0.001$).}
\label{table_twitter_tfidf_correlation}
\begin{tabular}{|p{1.4cm}|p{0.4cm}|p{1.7cm}|p{0.4cm}|p{1.3cm}|p{0.4cm}|}
\hline
\multicolumn{2}{| c |}{\textbf{Buddhist\%}} & \multicolumn{2}{ c |}{\textbf{(High) Price}} & \multicolumn{2}{ c |}{\textbf{Deprivation}} \\
\hline
Term & $\mathbf{\rho}$ & Term & $\mathbf{\rho}$ & Term & $\mathbf{\rho}$ \\
\hline
\textbf{think} & 0.44 & \textbf{luxury} & 0.66 & \textbf{east} & 0.39 \\
\textbf{long} & 0.42 & \textbf{tea} & 0.64 & \textbf{eastlondon} & 0.36 \\
rainy & 0.41 & \textbf{teatime} & 0.61 & \textbf{eastend} & 0.36 \\
\textbf{learn} & 0.40 & \textbf{delight} & 0.60 & \textbf{yeah} & 0.33 \\
presentation & 0.40 &\textbf{truffle} & 0.60 & \textbf{studio} & 0.33 \\
\textbf{mind} & 0.40 & \textbf{car} & 0.60 & \textbf{shit} & 0.32 \\
para & 0.40 & \textbf{classy} & 0.59 & \textbf{craftbeer} & 0.30 \\
todo & 0.40 & \textbf{stylish} & 0.59 & \textbf{ass} & 0.30 \\
thing & 0.40 & \textbf{gorgeous} & 0.59 & \textbf{music} & 0.30 \\
\textbf{heart} & 0.40 & \textbf{interiordesign} & 0.58 & neighbour & 0.29\\
\hline
\end{tabular}
\end{table}

We first examine Table~\ref{table_ya_tfidf_correlation} and provide examples of semantic similarity between \YA terms and the selected attributes. Words highlighted in bold are, in our view, the ones most associated with their respective attribute. For the attribute \textit{Deprivation}, the majority of the terms seem to be linked to issues of deprived areas. \quotes{Poverty}, \quotes{drug}, \quotes{victim}, all refer to social issues. \quotes{Rundown} and \quotes{slum} may be associated with the degradation of the surrounding environment. \quotes{Cockney} is a dialect traditionally spoken by working class, and thus less advantaged, Londoners. For the attribute (High) \textit{Price}, most terms seem to be related to aspects of places which may offer more expensive housing. Terms such as \quotes{fortune}, \quotes{diplomat}, and \quotes{aristocratic} are often associated with wealth. Others seem to reflect a posh lifestyle and status symbol: \quotes{townhouse}, \quotes{exclusive}, \quotes{celeb}, \quotes{fashionable}, \quotes{desirable}. For the attribute \textit{Jewish\%}, most of the terms seem to reflect aspects of this religion or be linguistically associated with it (i.e., \quotes{Jew} and \quotes{Jewish}). \quotes{Matzo} and \quotes{Kosher} are associated with the traditional Jewish cuisine; the former is a type of flat-bread, the latter is a way of preparing food. The \quotes{ark} is a specific part of the synagogue which contains sacred texts.

We now examine Table~\ref{table_twitter_tfidf_correlation}. For the attribute \textit{Deprivation}, nine words out of ten seem to be linked to more deprived areas. \quotes{East}, \quotes{eastlondon}, and \quotes{eastend}, for example, provide geographical information on where deprivation is more concentrated in London (i.e., East End). Other terms seem to be related to the presence of younger generation of creatives and artists in more deprived neighbourhoods. \quotes{Yeah}, \quotes{shit}, \quotes{ass}, may all be jargons commonly used by this section of population. \quotes{Studio}, \quotes{craftbeer}, \quotes{music} may instead refer to their main activities and occupations. For what concerns (high) \quotes{Price}, all the terms seem to relate to aspects of expensive areas, e.g. \quotes{luxury}, \quotes{classy}, and \quotes{stylish}. \quotes{Tea}, \quotes{teatime}, \quotes{delight}, \quotes{truffle} seem to relate to social activities of the upper class. For the attribute \quotes{Buddhist\%}, five terms out of ten are, in our view, associated with neighbourhoods where the majority of people is Buddhist or practise Buddhism. These terms seems to relate to aspects of this religion, e.g. \quotes{think}, \quotes{learn}, \quotes{mind}, etc.   

Interestingly, terms extracted from \YA and Twitter seem to offer two different kinds of knowledge. On one side, terms extracted from \YA are more encyclopedic as they tend to offer definitions or renowned aspects for each attribute. \quotes{Jewish\%} is, for example, related to aspects of the Jewish culture such as \quotes{matzo}, \quotes{harmony}, and \quotes{kosher}. \quotes{Deprivation} is associated with social issues such as \quotes{poverty} and \quotes{drug}, but also with a degraded urban environment (e.g., \quotes{rundown}, \quotes{slum}). On the other, Twitter words provide a kind of knowledge more related to current sociocultural aspects. This is the case, for example, of the jargon associated with \quotes{Deprivation} (e.g., \quotes{yeah}, \quotes{shit}), or of the culinary habits related to \quotes{High Prices} (e.g., \quotes{tea}, \quotes{truffle}).

\begin{table*}[ht]
\small
\centering
\caption{Prediction results in terms of $\rho$ using \YA and Twitter data. Results are averaged over $10$ folds and standard deviations are shown in parenthesis. Correlations are statistically significant (p-value $< 0.01$). Terms with the highest coefficients in regressions models are also provided.}
\label{table_ya_twitter_rgression_corr}
\begin{tabular}{ l | l | l | l | l}
 & \multicolumn{2}{l | }{\textbf{\YA}} & \multicolumn{2}{l}{\textbf{Twitter}}\\ 
 \hline
\textbf{Attribute}  & $\rho$ & Terms & $\rho$ & Terms \\
  \hline
Muslim $\%$ & $0.51 (0.07)$ & \textit{asian, barber} & $\mathbf{0.54} (0.05)$ & \textit{mileend, eastlondon} \\ 
Jewish $\%$ & $\mathbf{0.42} (0.08)$ & \textit{jewish, arab}& $0.13 (0.06)$ &\textit{rsa, rugby}\\ 
Hindu $\%$ & $0.32 (0.10)$ & \textit{stadium, cemetery} & $\mathbf{0.46} (0.09)$ & \textit{smokeyeye,asianbride} \\ 
Buddhist $\%$ & $0.24 (0.10)$ &\textit{minister, tourist}& $\mathbf{0.44} (0.07)$& \textit{theatre, prayforparis} \\ 
\hline
Black $\%$ & $\mathbf{0.60} (0.07)$ & \textit{gang, drug} & $0.44 (0.08)$ & \textit{southlondon, frank} \\ 
Asian $\%$ & $\mathbf{0.40} (0.07)$ & \textit{asian, barber} & $0.39 (0.05)$ & \textit{mileend, gymselfie} \\ 
White $\%$ & $\mathbf{0.58} (0.06)$ & \textit{essex, suburbia} & $0.45 (0.08)$ & \textit{essex, golf} \\ 
\hline
House Price & $\mathbf{0.69} (0.05)$ & \textit{money, compliment} & $0.68 (0.04)$ & \textit{dailyspecial, personaltrainer} \\ 
IMD & $\mathbf{0.69} (0.03)$ & \textit{notorious, cockney} & $0.56 (0.04)$ & \textit{np, eastlondon} \\ 
\end{tabular}
\end{table*}

\subsection{Prediction}
The results of the regression tasks performed over the selected set of demographic attributes, in terms of Pearson correlation coefficient ($\rho$), are presented in Table~\ref{table_twitter_tfidf_correlation}. Results are averaged over $10$ folds and standard deviations are displayed in parenthesis.

We can see that on average, performances of \YA and Twitter are very similarly with \YA having a slightly higher performance ($4\%$). Twitter data can predict the majority of the religion-related attributes with a higher correlation coefficient with the exception of population of \textit{Jewish}\%. On the other hand, \YA is superior to Twitter when predicting ethnicity related attributes such as population of \textit{White}\% and \textit{Black}\%. We have seen in Table~\ref{table_ya_twitter_tfidf_correlation_counts} that Twitter has very few correlated terms with the attributes \textit{White} (0) and \textit{Black} (2). 

We also observe that \textit{IMD} and \textit{Price} can be predicted with a high correlation coefficient using both \YA and Twitter. This can be due to the fact that there are many words in our dataset that can be related to the deprivation of a neighbourhood or to how expensive a neighbourhood is. This is also evident in Table~\ref{table_ya_twitter_tfidf_correlation_counts} where the number of correlated terms from both \YA and Twitter with these attributes are very high. On the other hand, terms that describe a religion or an ethnicity are more specific and lower in frequency. Therefore attributes that are related to religion or ethnicity are predicted with a lower accuracy. 

Table~\ref{table_ya_twitter_rgression_corr} further shows two terms that have the highest coefficients in the regressions models (across the majority of folds) for each attribute and source in the column \textit{Terms}. These terms are among the strong predictors of their respective attribute. Many of these terms appear to be related to the given demographic attribute (for both Twitter and \YA) and are also often amongst the top correlated terms presented in Tables~\ref{table_ya_tfidf_correlation} and ~\ref{table_twitter_tfidf_correlation}. We follow with some examples. According to the regression coefficients for the attribute \textit{Muslim\%}, neighbourhoods inhabited by a Muslim majority may be located in Mile End, an East London district  (i.e., Twitter terms \quotes{mileend} and \quotes{eastlondon}), see the presence of Asian population and barber shops (i.e., \YA terms \quotes{asian} and \quotes{barber}). According to the terms for \textit{Black\%}, neighbourhoods with a black majority tend to be located in the southern part of London (i.e., Twitter term \quotes{southlondon}) and experience social issues such as presence of criminal groups and drug use (i.e., \YA terms \quotes{gang} and \quotes{drug}). According to the terms for \textit{IMD}, more deprived areas seem to be located in the East End of London (i.e., Twitter term \quotes{eastlondon}) where the Cockney dialect is dominant (i.e., \YA term \quotes{cockney}). \YA and Twitter seem to complement one another in terms of information they provide through the terms associated with each attribute which in most cases are different. One noticeable difference is that Twitter tends to offer geographical information (e.g., \quotes{mileend}, \quotes{southlondon}, \quotes{essex}). On the other hand, terms from \YA sometimes match the name of the attribute (i.e. \quotes{asian} and \quotes{Jewish}).

In the Appendix, in Tables~\ref{table_all_attr1} and~\ref{table_all_attr2}, we show the prediction results for a wide range of $62$ demographic attributes using \YA and Twitter. For each attribute, we display two terms with the highest coefficient common between the majority of the folds. Attributes are divided into categories such as \textit{Religion}, \textit{Employment}, \textit{Education}, etc. Overall, the results show that \YA performs slightly better than Twitter with an average $1\%$ increase over all the attributes. Wilxocon signed rank test shows that their results are significantly different from each other (p-value $ < 0.01$). Outcomes in these tables show that on average, a wide range of demographic attributes of the population of neighbourhoods can be predicted using both \YA and Twitter with high performances of $0.54$ and $0.53$ respectively. While \YA outperforms Twitter in predicting attributes related to \textit{Ethnicity} and \textit{Employment}, Twitter performs better when predicting attributes relating to the \textit{Age Group}, and \textit{Car Ownership}. 

\section{Related Work}
The availability of a huge amount of data from many social media platforms has inspired researchers to study the relation between the data on these platforms and many real-world attributes. Twitter data, in particular, has been widely used as a social media source to make predictions in many domains. For example, box-office revenues are predicted using text from Twitter microblogs~\cite{asur2010predicting}. Prediction results have been predicted by performing content analysis on tweets~\cite{tumasjan2010predicting}. It is shown that correlations exist between mood states of the collective tweets to the value of Dow Jones Industrial Average (DJIA)~\cite{bollen2011twitter}. 

Predicting demographics of individual users using their language on social media platforms, especially Twitter, has been the focus of many research. Text from blogs, telephone conversations, and forum posts are utilised for predicting author's age~\cite{nguyen2011author} with a Pearson's correlation of $0.7$. Geo-tagged Twitter data have been used to predict the demographic information of authors such as first language, race, and ethnicity with correlations up to $~0.3$~\cite{eisenstein2011discovering}.

One aspect of urban area life that has been the focus of many research work in urban data mining is finding correlations between different sources of data and the deprivation index (IMD), of neighbourhoods across a city or a country~\cite{lathia2012hidden,quercia2012tracking}. Cellular data~\cite{smith2014poverty} and the elements present in an urban area~\cite{venerandi2015measuring} are among non-textual data sources that are shown to have correlations with a deprivation index. Also, flow of public transport data has been used to find correlations (with a correlation coefficient of $r = 0.21$) with IMD of urban areas available in UK census~\cite{lathia2012hidden}. Research shows that correlations of $r = 0.35$ exists between the sentiment expressed in tweets of users in a community and the deprivation index of the community~\cite{quercia2012tracking}.

Social media data has been used in many domains to find links to the real-world attributes. Data generated on QA platforms, however, has not been used in the past for predicting such attributes. In this paper, we use discussions on \YA QA platform to make predictions of demographic attribute of city neighbourhoods. Previous work in this domain has mainly focused on predicting the deprivation index of areas~\cite{quercia2012tracking}. In this work, we look at a wide range of attributes and report prediction results on $62$ demographic attributes. Additionally, work in urban prediction uses geolocation-based platforms such as Twitter. QA data that has been utilised in this paper does not include geolocation information. Utilising such data presents its own challenges.  

\section{Discussion}
In this paper, we investigate predicting values for real-world entities such as demographic attributes of neighbourhoods using discussions from QA platforms. We show that these attributes can be predicted using text features based on \YA discussions about neighbourhoods with a slightly higher correlation coefficient than predictions made using Twitter data.

\subsection{Limitations}
Here, we present some of the limitations of our work.

\paragraph{Unification of the units of analysis.} To unify the units of analysis, we take a heuristic approach. We do not cross-validate our results with other approaches. This is because of the lack of work in using non-geotagged text for predicting attributes of neighbourhoods in the current literature.
\paragraph{Coverage.} Our experiments in this paper is limited to the city of London. London is a cosmopolitan city and a popular destination for travellers and settlers. Therefore, many discussions can be found on \YA regarding its neighbourhoods. The coverage of discussions on QA platforms may not be sufficient for all cities of interest.

\bibliographystyle{aaai}
\bibliography{qa}
\newpage
\appendix
\begin{table*}[h]
\small
\centering
\caption{Prediction results for different categories and attributes in terms of $\rho$ using \YA and Twitter data. Results are averaged over $10$ folds and standard deviations are shown in parenthesis. All correlations are statistically significant with a p-value $< 0.01$. For each category, the difference in performance between the two sources are highlighted in the column related to the outperforming (i.e. upward arrow) source. }
\label{table_all_attr1}
\begin{tabular}{ l | r | l | r | l }
 & \multicolumn{2}{l | }{\textbf{\YA}} & \multicolumn{2}{l}{\textbf{Twitter}}\\ 
 \hline
\textbf{Attribute}  & $\rho$ & Terms & $\rho$ & Terms \\
\hline 
\hline
\multicolumn{1}{ c |}{\textbf{Price \& Deprivation}} & \textbf{0.69}& \textbf{5} \%$\uparrow$ &0.64& \\
\hline
House Price & \textbf{0.69} & \textit{money, compliment} & 0.68 & \textit{dailyspecial, personaltrainer}\\ 
IMD & \textbf{0.69} & \textit{notorious, cockney} & 0.56 & \textit{np, eastlondon}\\ 
\hline
\multicolumn{1}{ c |}{\textbf{Religion}} & 0.37& &\textbf{0.39}& \textbf{2} \%$\uparrow$\\
\hline
Muslim \% & 0.51 & \textit{asian, barber}& \textbf{0.54}& \textit{mileend, eastlondon}\\ 
Jewish \% & \textbf{0.42}& \textit{jewish, arab} & 0.13& \textit{rsa, rugby}\\ 
Hindu \% & 0.32 & \textit{stadium, cemetery}& \textbf{0.46}& \textit{smokeyeye,asianbride}\\ 
Buddhist \% & 0.24& \textit{minister, tourist} & \textbf{0.44}& \textit{theatre, prayforparis}\\ 
\hline
\multicolumn{1}{ c |}{\textbf{Ethnicity}} & \textbf{0.49}& \textbf{6} \%$\uparrow$ &0.43& \\
\hline
Black \% & \textbf{0.6} & \textit{gang, drug}& 0.44& \textit{southlondon,frank}\\ 
Asian \% & \textbf{0.40} & \textit{asian, barber}& 0.39& \textit{mileend, gymselfie}\\ 
White \% & \textbf{0.58}& \textit{essex, suburbia} & 0.45& \textit{essex, golf}\\ 
Mixed \% & 0.37 & \textit{reggae,gang} &\textbf{0.45}& \textit{studio,southlondon}\\ 
\hline
\multicolumn{1}{ c |}{\textbf{Age Group}} & 0.56 & & \textbf{0.60} & \textbf{4} \%$\uparrow$\\
\hline
0-15 & 0.53 & \textit{mortgage,crappy}& \textbf{0.54}& \textit{uel,ikea}\\ 
16-29 & \textbf{0.66}& \textit{student,music} & \textbf{0.66}& \textit{drum,campus}\\ 
30-44 & 0.5 & \textit{cycle,psychic}& \textbf{0.6}& \textit{nffc,loyaltylunch}\\ 
45-64 & 0.46 & \textit{temporarily,underrate}& \textbf{0.57}& \textit{essex,golf}\\ 
65 Plus & \textbf{0.62}& \textit{hospital,outskirts} & \textbf{0.62}& \textit{golf,thearcher}\\ 
Working Age & 0.58 & \textit{foody,triple}& \textbf{0.64}& \textit{ukjob,delay}\\ 
\hline
\multicolumn{1}{ c |}{\textbf{Household Composition}} & 0.55 & &\textbf{ 0.58} & \textbf{3} \%$\uparrow$\\
\hline
Couple With Dependent Children\% & 0.57& \textit{belt,affordability} & \textbf{0.76}& \textit{blondieblue,xoxo}\\ 
Couple Without Dependent Children\% & \textbf{0.59} & \textit{role,essex}& 0.55& \textit{essex,semipermanentmakeup}\\ 
Lone Parent Household\% & \textbf{0.61} & \textit{gang,mortgage}& 0.38& \textit{helpme,ikea}\\ 
One Person Household\% & 0.55 & \textit{hotel,fashionable}& \textbf{0.7}& \textit{personaltrainer,wine}\\ 
At Least One Person 16+ English 1st Language\% & 0.52 & \textit{essex,outskirts}& \textbf{0.55}& \textit{golf,essex}\\ 
No People Aged 16+  English 1st Language\% & 0.46 & \textit{asian,foreigner}& \textbf{0.53}& \textit{tube,edgwareroad}\\ 
\hline
\multicolumn{1}{ c |}{\textbf{Residential Status}} & 0.58 & & \textbf{0.63} & \textbf{5} \%$\uparrow$\\
\hline
Owned Outright \% & \textbf{0.69} & \textit{chelmsford,outskirts}& 0.61& \textit{starbucks,grand}\\ 
Owned With A Mortgage Or Loan \% & 0.67 & \textit{belt,scummy}& \textbf{0.74}& \textit{barbergang,essex}\\ 
Social Rented \% & \textbf{0.65} & \textit{cockney,dump}& 0.56& \textit{ikea,studio}\\ 
Private Rented \% & 0.48& \textit{hotel,privacy} & \textbf{0.6}& \textit{tube,edgwareroad}\\ 
Household One$+$ Usual Resident \% & 0.49 & \textit{mortgage,gang}& \textbf{0.59}& \textit{eastlondon,londonbridge}\\ 
Household No Usual Residents \% & 0.39 & \textit{hotel,square}& \textbf{0.61}& \textit{hotel,marblearch}\\ 
Whole House Or Detached \% & 0.56 & \textit{underrate,retiree}& \textbf{0.69}& \textit{instafamily,crochet}\\ 
Whole House Or Semi Detached \% & 0.65& \textit{benefit,suburbia} & \textbf{0.72}& \textit{essex,semipermanentmakeup}\\ 
Whole House Or Terraced \% & \textbf{0.55} & \textit{cypriot,value}& 0.53& \textit{followforfollow,brockley}\\ 
Flat Or Apartment \% & 0.68 & \textit{location,inexpensive}& \textbf{0.76}& \textit{nffc,pcm}\\ 
Sale & \textbf{0.53} & \textit{commute,upmarket}& 0.5& \textit{personaltrainer,crochet}\\ 
  \hline
\multicolumn{1}{ c |}{\textbf{Employment}} & \textbf{0.54} & \textbf{8} \%$\uparrow$ & 0.46&\\
\hline
No Adults Employed - Dependent Children \% & \textbf{0.52} & \textit{interchange,cockney}& 0.33& \textit{ikea,gymtime}\\ 
Lone Parent Not In Employment Percent & \textbf{0.55} & \textit{slum,cockney}& 0.53& \textit{mileend,edgwareroad}\\ 
Economically Active Total & 0.46& \textit{suite,deprive} & \textbf{0.57}& \textit{railway,kensalrise}\\ 
Economically Inactive Total & \textbf{0.58}& \textit{student,triple} & \textbf{0.58}& \textit{mileend,gymtime}\\ 
Economically Active Employee & 0.28 & \textit{cycle,deprive}& \textbf{0.4}& \textit{railway,royaltylunch}\\ 
Economically Active Self Employed & \textbf{0.65} & \textit{jewish,affordability}& 0.47& \textit{rugby,northlondon}\\ 
Economically Active Unemployed & \textbf{0.66} & \textit{cockney,drug}& 0.5& \textit{np,eastlondon}\\ 
Economically Active Full Time Student & \textbf{0.57} & \textit{student,asian}& 0.49& \textit{np,instrumental}\\ 
Employment Rate & \textbf{0.54} & \textit{commute,suburban}& 0.36& \textit{railway,barbergang}\\ 
Unemployment Rate & \textbf{0.59} & \textit{notorious,cockney}& 0.40& \textit{swap,eastlondon}\\ 
\end{tabular}
\end{table*}

\begin{table*}[ht]
\small
\centering
\caption{cont.}
\label{table_all_attr2}
 \begin{tabular}{ l | r | l | r | l }
& \multicolumn{2}{l | }{\textbf{\YA}} & \multicolumn{2}{l}{\textbf{Twitter}}\\ 
 \hline
\textbf{Attribute}  & $\rho$ & Terms & $\rho$ & Terms \\
\hline
\hline
\multicolumn{1}{ c |}{\textbf{Education}} & 0.54 & & \textbf{0.58} & \textbf{4}\%$\uparrow$\\
\hline
No Qualifications & \textbf{0.62} & \textit{scummy,cockney}& 0.55& \textit{eastlondon,puregym}\\ 
Highest Level Of Qualification Level 1 \% & 0.68& \textit{essex,scummy} & \textbf{0.72}& \textit{eastlondon,hackneywick}\\ 
Highest Level Of Qualification Level 2 \% & 0.69 & \textit{scummy,role}& \textbf{0.77}& \textit{followforfollow,tattoo}\\ 
Highest Level Of Qualification Apprenticeship \% & 0.56 & \textit{role,truck}& \textbf{0.75}& \textit{tatemodern,oldstreet}\\ 
Highest Level Of Qualification Level 3 \% & 0.16 & \textit{role,fish}& \textbf{0.23}& \textit{bttower,fresher}\\ 
Highest Level Of Qualification Level 4 \% And Above & \textbf{0.71} & \textit{scholarship,affordability}& 0.62& \textit{rugby,cave}\\ 
Highest Level Of Qualification Other \% & 0.38 & \textit{employer,stadium}& \textbf{0.39}& \textit{endorphin,tube}\\ 
Schoolchildren And Full Time Students 18$+$ \% & 0.53 & \textit{student,chips}& \textbf{0.59}& \textit{tube,np}\\ 
  \hline
  \multicolumn{1}{ c |}{\textbf{Health}} & 0.42 & & 0.42 & \\
  \hline
Day To Day Activities Limited A Lot \% & \textbf{0.33} & \textit{cockney,gang}& 0.31& \textit{eastlondon,shisha}\\ 
Day To Day Activities Limited A Little \% & 0.4 & \textit{gloom,puppy}& \textbf{0.52}& \textit{cafc,hackneywick}\\ 
Day To Day Activities Not Limited \% & \textbf{0.39}& \textit{commute,park} & 0.37& \textit{tea,enjoysmilelive}\\ 
Very Good Or Good Health \% & \textbf{0.48} & \textit{commute,park}& 0.39& \textit{rwc,tea}\\ 
Fair Health \% & 0.52 & \textit{scummy,gang}& \textbf{0.59}& \textit{streetfood,coy}\\ 
Bad Or Very Bad Health \% & \textbf{0.35} & \textit{cockney,gang}& 0.34& \textit{eastlondon,east}\\ 
\hline
  \multicolumn{1}{ c |}{\textbf{Car Ownership}} & 0.62 & & \textbf{0.71}&\textbf{9}\%$\uparrow$\\
  \hline
No Cars Or Vans In Household \% & \textbf{0.72}& \textit{brewery,cockney} & 0.71& \textit{tube,groove}\\ 
1 Car Or Van In Household \% & 0.67 & \textit{suburban,grounds}& \textbf{0.7}& \textit{onelife,supercar}\\ 
2 Cars Or Vans In Household \% & 0.67& \textit{hospital,outskirts} & \textbf{0.71}& \textit{thearcher,golf}\\ 
3 Cars Or Vans In Household \% & 0.57& \textit{role,belt} & \textbf{0.75}& \textit{dailypic,boxpark}\\ 
4 Or More Cars Or Vans In Household \% & 0.49 & \textit{freehold,residential}& \textbf{0.69}& \textit{rugby,cave}\\
\hline
\hline
Average & $\mathbf{0.54}$& & 0.53& \\ \hline
  \end{tabular}
\end{table*}

\end{document}